\documentclass[10pt,twocolumn,letterpaper]{article}

\usepackage[pagenumbers]{cvpr} 

\usepackage{xcolor}
\newcommand\gray[1]{{\textcolor{gray}{#1}}}

\usepackage{adjustbox}
\usepackage{multirow}
\usepackage{bm}

\definecolor{cvprblue}{rgb}{0.21,0.49,0.74}
\usepackage[pagebackref,breaklinks,colorlinks,citecolor=cvprblue]{hyperref}

\title{ECLIPSE: Efficient Continual Learning in Panoptic Segmentation \\with Visual Prompt Tuning}

\author{
Beomyoung Kim$^{1,2}$\hspace{1.5em}Joonsang Yu$^{1}$\hspace{1.5em}Sung Ju Hwang$^{2}$\\ \\
{NAVER Cloud, ImageVision$^1$\hspace{3em}KAIST$^2$}\\
{\tt\small \{beomyoung.kim,joonsang.yu\}@navercorp.com, sjhwang82@kaist.ac.kr}
}

\begin{document}
\maketitle

\begin{abstract}
Panoptic segmentation, combining semantic and instance segmentation, stands as a cutting-edge computer vision task. 
Despite recent progress with deep learning models, the dynamic nature of real-world applications necessitates continual learning, where models adapt to new classes (plasticity) over time without forgetting old ones (catastrophic forgetting). 
Current continual segmentation methods often rely on distillation strategies like knowledge distillation and pseudo-labeling, which are effective but result in increased training complexity and computational overhead. 
In this paper, we introduce a novel and efficient method for continual panoptic segmentation based on Visual Prompt Tuning, dubbed ECLIPSE. 
Our approach involves freezing the base model parameters and fine-tuning only a small set of prompt embeddings, addressing both catastrophic forgetting and plasticity and significantly reducing the trainable parameters. 
To mitigate inherent challenges such as error propagation and semantic drift in continual segmentation, we propose logit manipulation to effectively leverage common knowledge across the classes.
Experiments on ADE20K continual panoptic segmentation benchmark demonstrate the superiority of ECLIPSE, notably its robustness against catastrophic forgetting and its reasonable plasticity, achieving a new state-of-the-art.
The code is available at \url{https://github.com/clovaai/ECLIPSE}.
\end{abstract}
\vspace{-2mm}
\section{Introduction}

Image segmentation is a fundamental computer vision task, and it involves dividing an image into meaningful segments to facilitate easier analysis. 
One of the most advanced forms of image segmentation is \emph{panoptic segmentation}, which merges semantic segmentation (categorizing pixels into set categories) with instance segmentation (identifying individual objects within categories). 
Recent panoptic segmentation studies have made significant progress by leveraging convolutional neural networks~\cite{(panoptic-fpn)kirillov2019panoptic,(solov2)wang2020solov2,(k-net)zhang2021k,(panoptic-deeplab)cheng2020panoptic} and transformer-based architectures~\cite{(maskformer)cheng2021per,(mask2former)cheng2022masked,(oneformer)jain2023oneformer,(mask-dino)li2023mask,(detr)carion2020end}.

\begin{figure}[t]
    \centering
    \begin{subfigure}[t]{\linewidth}
        \includegraphics[width=\linewidth]{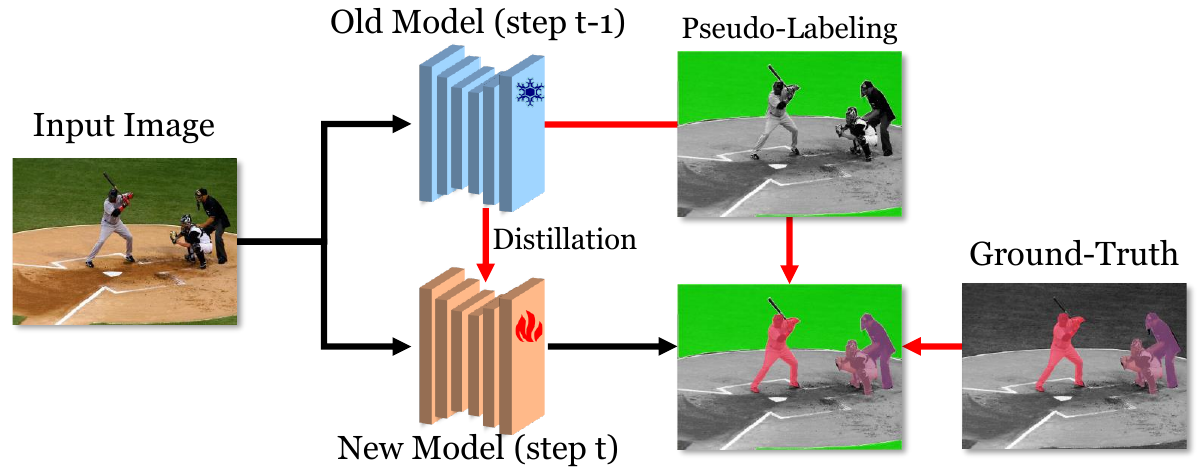}
        \vspace{-3mm}
        \caption{Previous methods}
        \label{fig:previous_method}
    \end{subfigure}
    \begin{subfigure}[t]{\linewidth}
        \includegraphics[width=\linewidth]{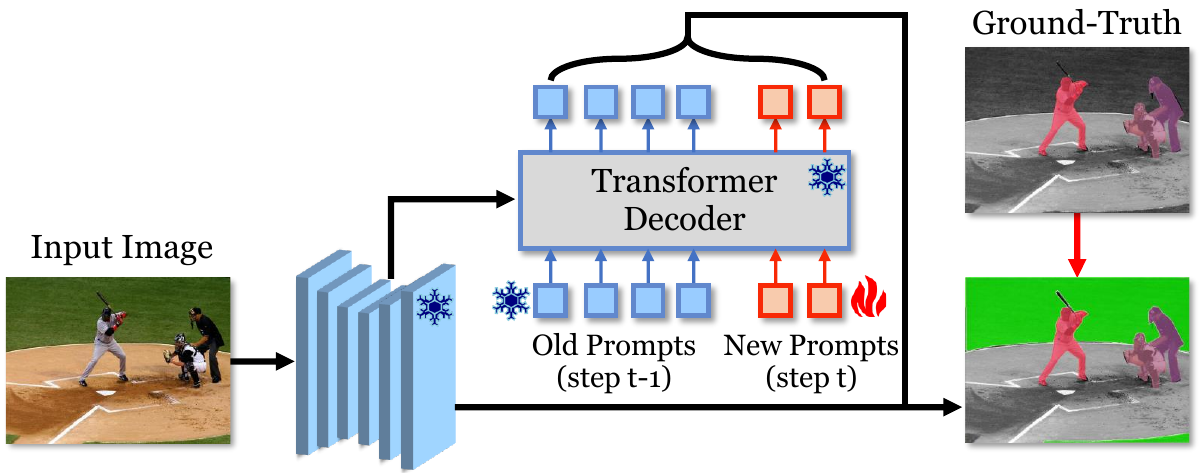}
        \vspace{-6mm}
        \caption{Our method}
        \label{fig:our_method}
    \end{subfigure}
    \caption{
        \textbf{Comparison of the overview of (a) previous methods and (b) our method.}
        Previous methods rely on distillation strategies such as knowledge distillation and pseudo-labeling, demanding more training complexity and computational overhead.
        In contrast, our method freezes all trained parameters and fine-tunes only a small set of prompt embeddings, robustly keeping the previous knowledge and extending the scalability of the model.
    }
    \vspace{-2mm}
\end{figure}
Despite these advances, the dynamic nature of the real world demands that models not only understand the present but also evolve over time. 
Continual image segmentation addresses this need by enabling models to learn new classes incrementally over time without forgetting the old classes.
It is critical in real-world applications where new classes emerge unpredictably, such as in robotics~\cite{(pnn)rusu2016progressive} and surveillance~\cite{(surveillance)ross2008incremental}. 
However, it is greatly challenging to preserve the previous class knowledge (avoiding \textbf{catastrophic forgetting}~\cite{french1999catastrophic}) and integrate new class information efficiently (\textbf{plasticity}) simultaneously.

Recently, various continual segmentation methods~\cite{(MiB)cermelli2020modeling,(PLOP)douillard2021plop,(RC)zhang2022representation,(SDR)michieli2021continual,(UCD)yang2022uncertainty,(ssul)cha2021ssul,(REMINDER)phan2022class,(RBC)zhao2022rbc,(incrementer)shang2023incrementer} have emerged, addressing the key challenges and showing notable improvements.
Most continual segmentation approaches often employ distillation strategies like knowledge distillation~\cite{(MiB)cermelli2020modeling,(PLOP)douillard2021plop,(RC)zhang2022representation} and pseudo-labeling~\cite{(PLOP)douillard2021plop,(ssul)cha2021ssul,(REMINDER)phan2022class}, as shown in Figure \ref{fig:previous_method}.
Knowledge distillation can alleviate catastrophic forgetting by transferring knowledge from an old model to a new model, and pseudo-labeling allows the new model to train with labels of the previously learned classes.
Though groundbreaking, these approaches involve trade-offs such as the need for doubled network forwarding and careful tuning of the hyperparameters ($e.g.,$ distillation loss weights and threshold for pseudo-labels), which increases the training complexity and computational overhead.
As the number of classes increases incrementally, maintaining a scalable and efficient distillation process can become challenging.
Moreover, while much of the research has focused on continual semantic segmentation, continual panoptic segmentation, which is more challenging in incorporating both semantic- and instance-level segmentation tasks, has been relatively underexplored.

In this paper, we propose a novel method, dubbed \textbf{ECLIPSE}, for \textbf{E}fficient \textbf{C}ontinual \textbf{L}earning \textbf{I}n \textbf{P}anoptic \textbf{SE}gmentation that leverages the potential of Visual Prompt Tuning (VPT)~\cite{(vpt)jia2022visual} and obviates the need for conventional distillation strategies.
Our approach begins with freezing all parameters of the base model and repeatedly fine-tunes a set of new prompt embeddings as new classes emerge, as shown in Figure \ref{fig:our_method}.
Our method inherently addresses catastrophic forgetting through model freezing and enhances plasticity via prompt tuning.
To the best of our knowledge, our approach is the first distillation-free continual panoptic segmentation, significantly reducing the trainable parameters and simplifying the continual segmentation process.

Despite these strengths, we confront inherent challenges in continual panoptic segmentation that necessitate further improvement.
Although model freezing preserves the prior knowledge, it can simultaneously propagate prior errors forward.
Moreover, the definition of \texttt{no-obj} class, which is required in inference to distinguish whether an output mask is no object or not, changes at each continual learning step, which is known as the semantic drift problem.
To circumvent the challenges, we propose a simple and effective strategy, called logit manipulation.
It allows the model to leverage the inter-class knowledge of all learned classes to more meaningfully manipulate the \texttt{no-obj} logit.
The dynamically updated \texttt{no-obj} logit helps suppress prior error predictions and mitigate the semantic drift issue at once.

Our comprehensive experiments on ADE20K~\cite{(ade20k)zhou2017scene} demonstrate that ECLIPSE achieves a new state-of-the-art in continual panoptic segmentation with a mere 1.3\% of the trainable parameters.
Notably, ECLIPSE shows outstanding robustness against catastrophic forgetting, especially as the number of continual steps increases, making a substantial improvement over existing methods.

In summary, the contributions of our paper are:
\begin{itemize}
    \item We successfully integrate VPT into continual panoptic segmentation, effectively mitigating catastrophic forgetting and efficiently extending the scalability of the model.
    \item We propose an effective logit manipulation strategy that circumvents the inherent challenges in continual panoptic segmentation: error propagation and semantic drift.
    \item We achieve state-of-the-art results on ADE20K with a significantly few number of trainable parameters.
\end{itemize}
    
\begin{figure*}[t]
    \centering
    \includegraphics[width=\linewidth]{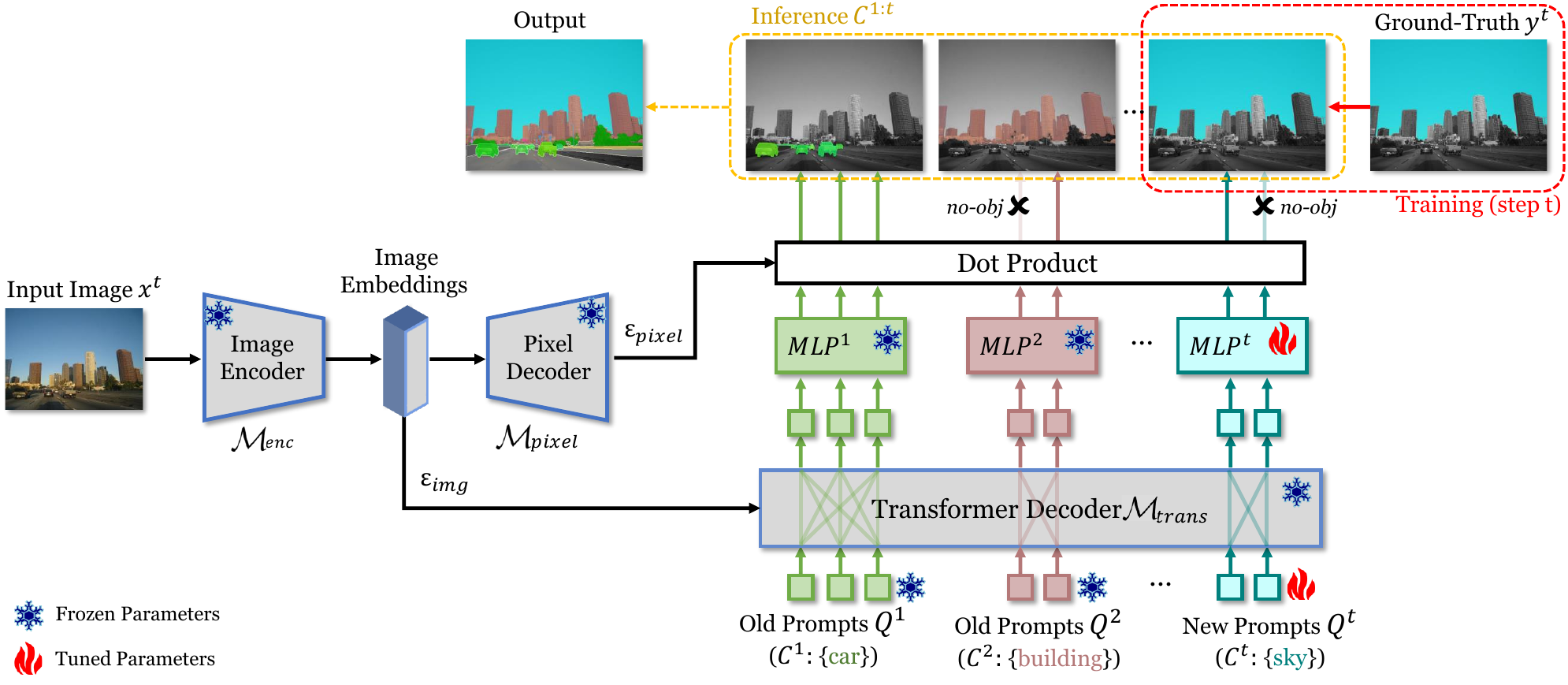}
    \caption{\textbf{Overview of ECLIPSE}. We freeze all trained parameters and fine-tune only a set of prompt embeddings $\mathbf{Q}^{t}$ alongside MLP layers to recognize a set of classes $\mathcal{C}^{t}$. In inference, we aggregate outputs from all prompt sets $\mathbf{Q}^{1:t}$ to segment all learned classes $\mathcal{C}^{1:t}$.}
    \label{fig:overview}
\end{figure*}

\section{Related Work}

\paragraph{Panoptic Segmentation}
is a cutting-edge task in computer vision, blending the concepts of semantic and instance segmentation to provide a comprehensive understanding of both `stuff' (amorphous regions like sky or grass) and `things' (countable objects like cars or people).
Pioneering works~\cite{(ps)kirillov2019panoptic,(panoptic-fpn)kirillov2019panoptic,(panoptic-deeplab)cheng2020panoptic} integrated semantic and instance segmentation tasks into a unified framework.
Following this, some methods~\cite{(solov2)wang2020solov2,(k-net)zhang2021k,(panoptic-fcn)li2021fully} introduced significant improvements by employing dynamic convolutions in a fully convolutional paradigm.
More recently, transformer-based architectures~\cite{(maskformer)cheng2021per,(mask2former)cheng2022masked,(mask-dino)li2023mask} have further advanced the field by leveraging the power of attention mechanisms. 
However, the challenge of continual panoptic segmentation, particularly in dynamically adapting to new classes without forgetting the old ones, remains a frontier for ongoing research.

\vspace{-4mm}
\paragraph{Continual Segmentation.}
To address the dynamic nature of real-world applications, continual segmentation emerges as an advanced task.
Pioneering work~\cite{(MiB)cermelli2020modeling} has discovered the distinct challenge in continual segmentation tasks, semantic drift, caused by \textit{background} class.
Most approaches~\cite{(MiB)cermelli2020modeling,(PLOP)douillard2021plop,(ssul)cha2021ssul,(SDR)michieli2021continual,(RC)zhang2022representation,(RBC)zhao2022rbc,(REMINDER)phan2022class} utilize distillation strategies such as knowledge distillation and pseudo-labeling to mitigate the semantic drift issue.
More recently, Incrementer~\cite{(incrementer)shang2023incrementer} leverages the architectural advantages of a transformer-based model using incremented class embeddings and multiple distillation strategies.
However, most of the research has focused on continual semantic segmentation, and continual panoptic segmentation, which is a more challenging task, is less explored.
CoMFormer~\cite{(comformer)cermelli2023comformer} is a pioneer work in continual panoptic segmentation using the universal segmentation model ($i.e.,$ Mask2Former~\cite{(mask2former)cheng2022masked}) to perform both panoptic and semantic segmentation tasks with query-based distillation strategy.
However, such distillation-based approaches increase the training complexity and computational overhead and demand careful tuning of hyperparameters such as loss weights and temperatures of distillation and threshold for pseudo-labeling.
Unlike them, our method is the first distillation-free approach for both panoptic and semantic segmentation tasks, simplifying the continual learning process and reducing training computations.

\paragraph{Visual Prompt Tuning (VPT) in Continual Learning.}
VPT~\cite{(vpt)jia2022visual} introduced an efficient and effective fine-tuning method for vision transformer models. 
By freezing the pre-trained parameters, they fine-tuned only a set of learnable prompts and achieved remarkable performance.
In the continual image classification field, there are several attempts to utilize the VPT.
Namely, L2P~\cite{(L2P)wang2022learning} and DualPrompt~\cite{(dualprompts)wang2022dualprompt} freeze the pre-trained model and select the most relevant prompts from a prompts pool in a key-value mechanism.
They achieved noticeable performance and demonstrated the potential of leveraging VPT in continual image classification.
We are the first VPT-based continual segmentation method tailored to address several distinct challenges in continual panoptic segmentation.

\section{Preliminary}

\subsection{Problem Setting}
Panoptic segmentation is an advanced task that unifies semantic- and instance-level segmentation tasks.
Compared to semantic segmentation without discrimination of individual instances, panoptic segmentation necessitates a more comprehensive unified framework to identify both `things' ($e.g.,$ individual car) and `stuff' ($e.g.,$ sky or road).

This paper mainly focuses on continual learning in panoptic segmentation, following the same setting in \cite{(comformer)cermelli2023comformer}.
Over continually arriving tasks at timesteps $t=1,\ldots,T$, at each step $t$, a training dataset $\mathcal{D}^t$ is introduced which contains image-label pairs $(\bm x^{t},\bm{y}^{t})$, where $\bm x^t$ represents an image and $\bm{y}^t$ its corresponding segmentation label.
Here, $\bm{y}^{t}$ is labeled only for the set of current classes $\mathcal{C}^{t}$, and other classes (previous $\mathcal{C}^{1:t-1}$ and future $\mathcal{C}^{t+1:T}$ classes) are not accessible.
Once a task is completed, the model is expected to segment for all classes in $\mathcal{C}^{1:t}$ while preventing catastrophic forgetting for $\mathcal{C}^{1:t-1}$ and incrementally learning for $\mathcal{C}^{t}$ (plasticity).

\subsection{Network Architecture: Mask2Former}
We adopt Mask2Former~\cite{(mask2former)cheng2022masked} as our baseline architecture, which is a transformer-based model for universal segmentation tasks including panoptic, instance, and semantic segmentation.
Unlike the previous paradigm of per-pixel classification, Mask2Former directly predicts a set of masks including their classes, termed mask classification.
Mask2Former consists of three kinds of modules: image encoder $\mathcal{M}_{enc}$, pixel decoder $\mathcal{M}_{pixel}$ and transformer decoder $\mathcal{M}_{trans}$.
The image encoder extracts image embedding $\mathcal{E}_{img}$ from the input image, and the pixel decoder converts image embedding into per-pixel embedding $\mathcal{E}_{pixel}$.
The transformer decoder takes $N$ learnable queries and generates $N$ mask embeddings by self-attention and cross-attention with image embedding.
Here, each query takes charge of representing an object (or no object).
Finally, $N$ mask proposals are produced via a dot product between $N$ mask embedding and per-pixel embedding.
Moreover, the category of each mask is assigned using the MLP-based classifier. 
When a mask proposal is classified as \textit{no-obj} class, it is dropped during inference.

For simplicity, we express the model $\mathcal{M}$ as a function which takes an image $\mathbf{x} \in \mathbb{R}^{3 \times H \times W}$ and queries $\mathbf{Q} \in \mathbb{R}^{N \times D}$ as inputs and output masks $\mathbf{m} \in \mathbb{R}^{N \times H \times W}$ and class logits $\mathbf{s} \in \mathbb{R}^{N \times C}$: 
\begin{gather}
    (\mathbf{s}, \mathbf{m}) = \mathcal{M}(\mathbf{x}, \mathbf{Q}), \\
    \mathcal{M}(\mathbf{x}, \mathbf{Q}) = \text{MLP}(\mathcal{M}_{trans}(\mathcal{E}_{img}, \mathbf{Q})) \otimes \mathcal{E}_{pixel}, \\
    \mathcal{E}_{img} = \mathcal{M}_{enc}(\mathbf{x}), \text{   } \mathcal{E}_{pixel} = \mathcal{M}_{pixel}(\mathcal{E}_{img}),
\end{gather}
where $N$ is the number of queries, $D$ is the dimension for the query embedding, and $C$ is the number of classes.
\section{Method}

\subsection{Prompt Tuning for Continual Segmentation}
\label{method:prompt_tuning}
We present a novel approach, named ECLIPSE, for efficient continual panoptic segmentation which leverages Visual Prompt Tuning (VPT)~\cite{(vpt)jia2022visual} for the integration of new classes without the conventional distillation strategies.
Our approach begins with the initial training ($t{=}1$) of all parameters of the model $\mathcal{M}$ on the set of base classes $\mathcal{C}^{1}$:
\begin{gather}
    (\mathbf{s}^{1}, \mathbf{m}^{1}) = \textcolor{red}{\mathcal{M}}(\mathbf{x}, \textcolor{red}{\mathbf{Q}^{1}}).
\end{gather}

After this initial training, we apply a \textit{freeze-and-tune} strategy repeatedly.
Namely, when new classes are introduced ($t{>}1$), we freeze all trained parameters to preserve the acquired knowledge and fine-tune a set of learnable prompt embeddings $\mathbf{Q}^{t} \in \mathbb{R}^{N^{t} \times D}$ alongside unshared MLP layers to recognize new classes $\mathcal{C}^{t}$:
\begin{gather}
    (\mathbf{s}^{t}, \mathbf{m}^{t}) = \textcolor{cyan}{\mathcal{M}}(\mathbf{x}, \textcolor{red}{\mathbf{Q}^{t}}) \text{  } (t>1), \\
    \textcolor{cyan}{\mathcal{M}}(\mathbf{x}, \textcolor{red}{\mathbf{Q}^{t}}) = \textcolor{red}{\text{MLP}^{t}}(\textcolor{cyan}{\mathcal{M}_{trans}}(\mathcal{E}_{img}, \textcolor{red}{\mathbf{Q}^{t}})) \otimes \mathcal{E}_{pixel}, \\
    \mathcal{E}_{img} = \textcolor{cyan}{\mathcal{M}_{enc}}(\mathbf{x}), \text{   } \mathcal{E}_{pixel} = \textcolor{cyan}{\mathcal{M}_{pixel}}(\mathcal{E}_{img}),
\end{gather}
where $\textcolor{red}{\bullet}$ and $\textcolor{cyan}{\bullet}$ denote \textcolor{red}{trainable} and \textcolor{cyan}{frozen} parameters, respectively.
In short, we treat each set of prompts $\mathbf{Q}^{t}$ as a discrete task-specific module, solely devoted to the recognition of $\mathcal{C}^{t}$ classes.
As the continual steps progress, we stably extend the scalability of the model through the lightweight task-specific prompt sets.
During the inference phase, we aggregate outputs from all the prompt sets across steps. $\mathbf{Q}^{1:t}$, allowing the model to segment all learned classes $\mathcal{C}^{1:t}$:
\begin{gather}
    (\mathbf{s}^{1:t}, \mathbf{m}^{1:t}) = \mathcal{M}(\mathbf{x}, \mathbf{Q}^{1:t}).
\end{gather}
Our approach ensures that previous knowledge is preserved with model freezing, which prevents catastrophic forgetting, while also enabling the efficient integration of new knowledge through prompt tuning, enhancing the model's plasticity.

We introduce two prompt tuning strategies for continual panoptic segmentation, termed \textit{shallow} and \textit{deep}, inspired by \cite{(vpt)jia2022visual}.
The \textit{shallow} means tuning the prompt embeddings at the first transformer layer only, $\mathbf{Q}^{t}_{shallow}\in\mathbb{R}^{N^{t} \times D}$.
Whereas, the \textit{deep} denotes tuning the embeddings across all transformer layers, $\mathbf{Q}^{t}_{deep}{=}\{\mathbf{Q}^{t}_{1}, \mathbf{Q}^{t}_{2}, \cdots, \mathbf{Q}^{t}_{L} \}$ where $L$ is the number of transformer layers.
By default, we adopt the \textit{deep} prompt tuning and will present a detailed analysis in the experimental section.

\begin{figure}[t]
    \centering
    \includegraphics[width=\linewidth]{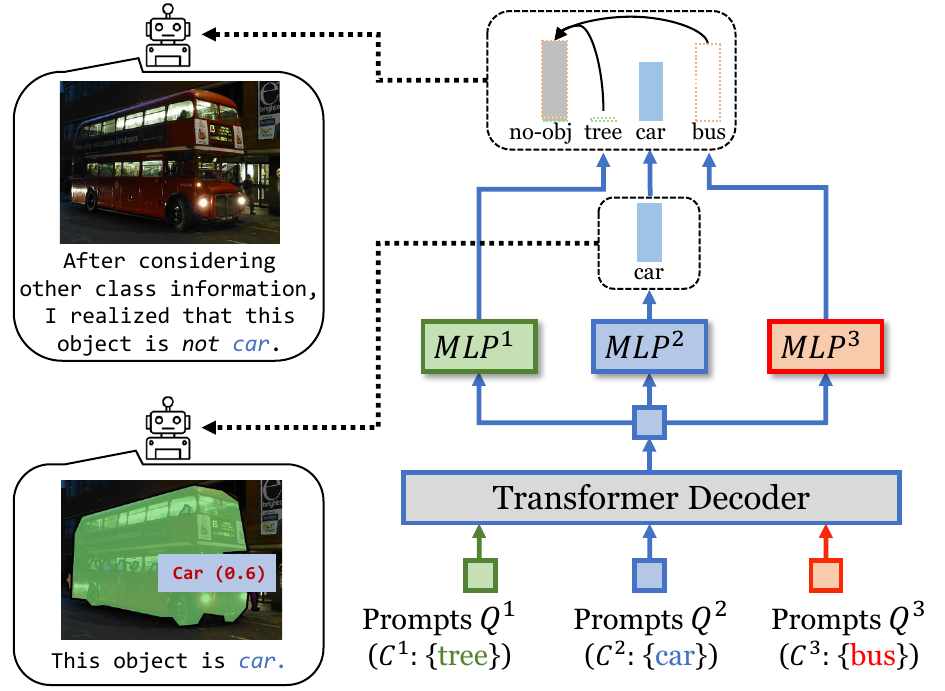}
    \caption{\textbf{Illustration of logit manipulation.} To alleviate semantic drift of \texttt{no-obj} class, we make a new \texttt{no-obj} logit leveraging the inter-class knowledge of all learned classes. Moreover, an erroneous prediction caused by semantic confusion of prior frozen parameters can be fixed through logit manipulation.
    }
    \label{fig:logit_manu}
\end{figure}

\subsection{Resolving Semantic Confusion and Drift}
Our method effectively addresses both catastrophic forgetting and plasticity in continual panoptic segmentation.
However, we confront an inherent issue, called error propagation. 
This issue arises because freezing the model, while helpful in preventing catastrophic forgetting, also means carrying prior errors forward.
These errors often originate from \textbf{semantic confusion}, where the model misclassifies objects due to their visual similarities with other classes. 
In a continual learning setting, semantic confusion becomes significant due to unawareness of future classes.
For example, a model that has learned to identify \texttt{car} might mistakenly recognize \texttt{bus} as \texttt{car} because it hasn’t learned to distinguish between them yet.

Moreover, there is a distinct issue in continual segmentation, called \textbf{semantic drift}, as discussed in prior works~\cite{(MiB)cermelli2020modeling,(PLOP)douillard2021plop,(ssul)cha2021ssul}.
Our baseline has a unique \texttt{no-obj} class with a corresponding MLP-classifier layer to identify whether an output mask is no object or not.
However, the definition of \texttt{no-obj} shifts with each continual step, including future classes not learned yet, past classes already learned, and the background.
The reliability of the \textit{no-obj} class is essential during inference, so semantic drift significantly impacts the performance of the model.

\begin{table*}[t]
  \centering
  \begin{subtable}{\linewidth}
      \centering
      \begin{adjustbox}{max width=\linewidth}
      \begin{tabular}{c|c|c|c|ccc|ccc|ccc}
        \toprule
        \multirow{2}{*}{Method} & \multirow{2}{*}{Backbone} & Trainable & \multirow{2}{*}{KD} & \multicolumn{3}{c|}{\textbf{100-5} (11 tasks)} & \multicolumn{3}{c|}{\textbf{100-10} (6 tasks)} & \multicolumn{3}{c}{\textbf{100-50} (2 tasks)} \\
         & & Params & & \textit{1-100} & \textit{101-150} & \textit{all} & \textit{1-100} & \textit{101-150} & \textit{all} & \textit{1-100} & \textit{101-150} & \textit{all} \\
        \midrule
        FT & R50 & 44.9M &  & 0.0 & 25.8 & 8.6 & 0.0 & 2.9 & 1.0 & 0.0 & 1.3 & 0.4 \\
        MiB~\cite{(MiB)cermelli2020modeling} & R50 & 44.9M & \checkmark & 24.0 & 6.5 & 18.1 & 27.1 & 10.0 & 21.4 & 35.1 & 19.3 & 29.8 \\
        PLOP~\cite{(PLOP)douillard2021plop} & R50 & 44.9M & \checkmark & 28.1 & 15.7 & 24.0 & 30.5 & 17.5 & 26.1 & 40.2 & 22.4 & 34.3 \\
        CoMFormer~\cite{(comformer)cermelli2023comformer} & R50 & 44.9M & \checkmark & 34.4 & 15.9 & 28.2 & 36.0 & 17.1 & 29.7 & 40.2 & 23.5 & 34.6 \\
        \midrule
        \multirow{2}{*}{\textbf{ECLIPSE}} & R50 & \textbf{0.60M} &  & \textbf{41.1} & \textbf{16.6} & \textbf{32.9} & \textbf{41.4} & \textbf{18.8} & \textbf{33.9} & \textbf{41.7} & \textbf{23.5} & \textbf{35.6}  \\
         & \gray{Swin-L} & \gray{0.60M} &  & \gray{48.0} & \gray{20.6} & \gray{38.9} & \gray{48.6} & \gray{25.5} & \gray{40.9} & \gray{48.2} & \gray{29.8} & \gray{42.0} \\
        \midrule
        joint & R50 & & & 43.2 & 32.1 & 39.5 & 43.2 & 32.1 & 39.5 & 43.2 & 32.1 & 39.5 \\
        \bottomrule
      \end{tabular}
      \end{adjustbox}
      \caption{}
      \vspace{2mm}
  \end{subtable}
  \begin{subtable}{\linewidth}
      \centering
      \begin{adjustbox}{max width=\linewidth}
      \begin{tabular}{c|c|c|c|ccc|ccc|ccc}
        \toprule
        \multirow{2}{*}{Method} & \multirow{2}{*}{Backbone} & Trainable & \multirow{2}{*}{KD} & \multicolumn{3}{c|}{\textbf{50-10} (11 tasks)} & \multicolumn{3}{c|}{\textbf{50-20} (6 tasks)} & \multicolumn{3}{c}{\textbf{50-50} (3 tasks)} \\
         & & Params & & \textit{1-50}~~ & ~~\textit{51-150} & \textit{all} & \textit{1-50}~~ & ~~\textit{51-150} & \textit{all} & \textit{1-50}~~ & ~~\textit{51-150} & \textit{all} \\
        \midrule
        FT & R50 & 44.9M &  & 0.0  & 1.7  & 1.1 & 0.0  & 4.4  & 2.9 & 0.0 & 12.0 & 8.1    \\
        MiB~\cite{(MiB)cermelli2020modeling} & R50 & 44.9M & \checkmark  & 34.9  & 7.7  & 16.8 & 38.8  & 10.9  & 20.2 & 42.4 & 15.5 & 24.4   \\
        PLOP~\cite{(PLOP)douillard2021plop} & R50 & 44.9M & \checkmark & 39.9 & 15.0  & 23.3 & 43.9  & 16.2  & 25.4 & 45.8 & 18.7 & 27.7    \\
        CoMFormer~\cite{(comformer)cermelli2023comformer} & R50 & 44.9M & \checkmark & 38.5 & 15.6 & 23.2 & 42.7 & 17.2 & 25.7 & 45.0 & 19.3 & 27.9  \\
        \midrule
        \multirow{2}{*}{\textbf{ECLIPSE}} & R50 & \textbf{0.60M} &  & \textbf{45.9} & \textbf{17.3} & \textbf{26.8} & \textbf{46.4} & \textbf{19.6} & \textbf{28.6} & \textbf{46.0} & \textbf{20.7} & \textbf{29.2}    \\
        & \gray{Swin-L} & \gray{0.60M} &  & \gray{52.8} & \gray{22.9} & \gray{32.9} & \gray{53.2} & \gray{25.7} & \gray{34.8} & \gray{53.0} & \gray{25.3} & \gray{34.5} \\
        \midrule
        joint & R50 & & & 50.2 & 34.1 & 39.5 & 50.2 & 34.1 & 39.5 & 50.2 & 34.1 & 39.5 \\
        \bottomrule
      \end{tabular}
      \end{adjustbox}
      \caption{}
      \vspace{-2mm}
  \end{subtable}
  \caption{
    \textbf{Continual Panoptic Segmentation} results on ADE20K dataset in PQ when the number of base classes $|\mathcal{C}^{1}|$ is (a) 100 and (b) 50. \textit{KD} denotes using distillation strategies, which demands more trainable parameters and computational overhead. All methods use the same network of Mask2Former~\cite{(mask2former)cheng2022masked} with ResNet-50~\cite{(resnet)he2016deep} backbone. \textit{joint} means an oracle setting training all classes offline at once.
  }
  \label{tab:comparision_adps}
\end{table*}

To simultaneously tackle the semantic confusion and drift issues, we propose a simple yet effective method, called \textbf{logit manipulation}.
First, we eliminate the MLP classifier for \texttt{no-obj} because the \texttt{no-obj} logit is no longer reliable due to semantic drift.
Instead, we generate a new \texttt{no-obj} logit by leveraging the mutual information of old and new classes to make \texttt{no-obj} information more meaningful.
For example, as shown in \figurename~\ref{fig:logit_manu}, the decoder output of prompt $\mathbf{Q}^2$ is fed into other $\textit{MLP}^1$ and $\textit{MLP}^3$ layers and then the \texttt{no-obj} logit is manipulated by aggregating logits from the $\textit{MLP}$ layers.
In our approach, as the outputs from $\mathbf{Q}^{t}$ are responsible for prediction only for $\mathbf{C}^{t}$ classes, logits of other classes that do not belong to $\mathbf{C}^{t}$ can be treated as \texttt{no-obj} class.
Given the set of prompts $\mathbf{Q}_{t}$ at step $t$, \texttt{no-obj} logits $s_{t}^{no-obj}$ are manipulated as:
\begin{gather}
    s^{\mathbf{C}^{1:T}}_{t} = \text{MLP}^{1:T}(\mathbf{Q}_{t}), \\
    s^{no-obj}_{t} = \delta \times (\sum_{k=1}^{t-1}s^{\mathbf{C}^{k}}_{t} + \sum_{k=t+1}^{T}s^{\mathbf{C}^{k}}_{t}), \\
    c_{t} = \text{argmax}(s^{no-obj}_{t}, s_{t}^{\mathbf{C}^{t}}),
\end{gather}
where $c_{t}$ is the class indexes of output masks $m_{t}$ and $\delta$ is a scalar hyperparameter for logit modulation.
The dynamically manipulated \texttt{no-obj} logit helps in suppressing the propagated erroneous predictions and inherently resolves semantic drift because the \texttt{no-obj} logit is meaningfully updated.
We note that the logit manipulation is applied only at the inference stage and $\delta$ is the post-processing hyperparameter.
Unlike our baseline employs \textit{softmax} activation (relative logits) in the classification, we use \textit{sigmoid} activation (independent logits) to leverage the distinct information associated with each class, inspired by \cite{(ssul)cha2021ssul}.

\begin{table*}[t]
  \centering
  \begin{adjustbox}{max width=\linewidth}
  \begin{tabular}{c|c|c|c|ccc|ccc|ccc}
    \toprule
    \multirow{2}{*}{Method} & \multirow{2}{*}{Backbone} & Trainable & \multirow{2}{*}{KD} & \multicolumn{3}{c|}{\textbf{100-5} (11 tasks)} & \multicolumn{3}{c|}{\textbf{100-10} (6 tasks)} & \multicolumn{3}{c}{\textbf{100-50} (2 tasks)} \\
     & & Params & & \textit{1-100} & \textit{101-150} & \textit{all} & \textit{1-100} & \textit{101-150} & \textit{all} & \textit{1-100} & \textit{101-150} & \textit{all} \\
    \midrule
    SDR$^{\dagger}$~\cite{(SDR)michieli2021continual} & R101 & 60.4M & \checkmark  & - & - & - & 28.9 & 7.4 & 21.7 & 37.4 & 24.8 & 33.2\\
    UCD$^{\dagger}$~\cite{(UCD)yang2022uncertainty} & R101 & 60.4M & \checkmark & - & - & - & 40.8 & 15.2 & 32.3 & 42.1 & 15.8 & 33.3  \\
    SPPA$^{\dagger}$~\cite{(SPPA)lin2022continual} & R101 & 60.4M & \checkmark & - & - & - & 41.0 & 12.5 & 31.5 & 42.9 & 19.9 & 35.2 \\
    RCIL$^{\dagger}$~\cite{(RC)zhang2022representation} & R101 & 58.0M & \checkmark & 38.5 & 11.5 & 29.6 & 39.3 & 17.6 & 32.1 & 42.3 & 18.8 & 34.5  \\
    SSUL$^{\dagger}$~\cite{(ssul)cha2021ssul} & R101 & 1.78M & \checkmark & 39.9 & 17.4 & 32.5 & 40.2 & 18.8 & 33.1 & 41.3 & 18.0 & 33.6  \\
    REMINDER$^{\dagger}$~\cite{(REMINDER)phan2022class} & R101 & 60.4M & \checkmark & - & - & - & 39.0 & 21.3 & 33.1 & 41.6 & 19.2 & 34.1 \\
    \midrule
    MiB~\cite{(MiB)cermelli2020modeling} & R101 & 63.4M & \checkmark & 21.0 & 6.1 & 16.1 & 23.5 & 10.6 & 26.6 & 37.0 & 24.1 & 32.6 \\
    PLOP~\cite{(PLOP)douillard2021plop} & R101 & 63.4M & \checkmark & 33.6 & 14.1 & 27.1 & 34.8 & 15.9 & 28.5 & 43.4 & 25.7 & 37.4 \\
    CoMFormer~\cite{(comformer)cermelli2023comformer} & R101 & 63.4M & \checkmark & 39.5 & 13.6 & 30.9  & 40.6 & 15.6 & 32.3 & 43.6 & 26.1 & 37.6 \\
    \midrule
    \textbf{ECLIPSE} & R101 & \textbf{0.60M} &  & \textbf{43.3} & \textbf{16.3} & \textbf{34.2} & \textbf{43.4} & \textbf{17.4} & \textbf{34.6} & \textbf{45.0} & \textbf{21.7} & \textbf{37.1}\\
    \midrule
    joint & R101 & & & 46.9 & 35.6 & 43.1 & 46.9 & 35.6 & 43.1 & 46.9 & 35.6 & 43.1  \\
    \bottomrule
  \end{tabular}
  \end{adjustbox}
  \caption{
    \textbf{Continual Semantic Segmentation} results on ADE20K dataset in mIoU.
    All methods use the same backbone network of ResNet-101~\cite{(resnet)he2016deep}.
    $\dagger$ denotes that using DeepLab-V3~\cite{(deeplabv3)chen2017rethinking} network, otherwise using Mask2Former~\cite{(mask2former)cheng2022masked} network architecture.
  }
  \label{tab:comparision_adss_100}
\end{table*}

\section{Experiments}

\subsection{Experimental Setting.}
\paragraph{Dataset and Evaluation Metrics.}
We conduct experiments on ADE20K~\cite{(ade20k)zhou2017scene} dataset that consists of 150 classes with 100 things and 50 stuff categories and provides both panoptic and semantic segmentation benchmarks.
Compared to COCO~\cite{(coco)lin2014microsoft} containing an average of 7.7 instances and 3.5 classes per image and VOC~\cite{(voc)everingham2010pascal} containing an average of 2.3 instances and 1.4 classes per image, ADE20K contains an average of 19.5 instances and 9.9 classes.
We adopt Panoptic Quality (PQ) for evaluating continual panoptic segmentation performance and mean Inter-over-Union (mIoU) for continual semantic segmentation.
In detail, the PQ is defined by recognition quality~(RQ) and segmentation quality~(SQ):
\begin{equation}
\footnotesize
\mathrm{PQ}=\underbrace{\frac{\sum_{(p, g) \in T P} \operatorname{IoU}(p, g)}{|T P|}}_{\text {segmentation quality(SQ) }} \times \underbrace{\frac{|T P|}{|T P|+\frac{1}{2}|F P|+\frac{1}{2}|F N|}}_{\text {recognition quality(RQ) }},
\end{equation}
where $\operatorname{IoU}(p, g)$ is the intersection-over-union between the predicted mask $p$ and the ground truth $g$, and $TP$, $FP$, and $FN$ denote true-positive, false-positive, and false-negative, respectively.
After the last continual step $T$, we report the performances for the base classes ($\mathcal{C}^{1}$), new classes ($\mathcal{C}^{2:T}$), and all classes ($\mathcal{C}^{1:T}$).

\paragraph{Incremental Protocol.} 
Following previous continual segmentation methods~\cite{(MiB)cermelli2020modeling, (comformer)cermelli2023comformer}, we construct numerous challenging incremental protocols, termed as \texttt{\small{(BASE CLASSES)}}-\texttt{\small{(NEW CLASSES)}}.
For instance, \texttt{100-10} scenario means firstly learning 100 base classes and incrementally learning 10 new classes 5 times ($T{=}6)$.
The larger number of continual steps implies a more challenging scenario.
The seminal work~\cite{(MiB)cermelli2020modeling} introduced two different settings, \textit{disjoint} and \textit{overlap}. Here, we mainly follow the \textit{overlap} setting that is more challenging and realistic and provide results with the \textit{disjoint} setting in our supplementary material.

\paragraph{Implementation Details.}
Our implementation is based on the previous continual panoptic segmentation method, CoMFormer~\cite{(comformer)cermelli2023comformer}.
Specifically, we implement our method on Mask2Former~\cite{(mask2former)cheng2022masked} codebase using the backbone ResNet-50~\cite{(resnet)he2016deep} for continual panoptic segmentation and ResNet-101 for continual semantic segmentation.
Moreover, we set the dimension of prompt embeddings $D$ to 256, the number of transformer layers $L$ to 9, and $\text{MLP}^{t}$ consists of 2 hidden layers of 256 channels.
We set the number of incremented prompts to the number of incremented classes, $N^{t}{=}|\mathcal{C}^{t}|$, and the minimum value of $N^{t}$ to 10 to handle images containing more than 10 objects in $\mathcal{C}^{t}$.
For objective functions, we employ dice and binary cross-entropy loss functions as mask loss and binary cross-entropy loss function as classification loss, after \textit{bipartite matching} between predictions and ground-truths.
We train the model for 1600 iterations per class with a learning rate of 0.0001 for the first step and 0.0005 for the following steps.

\begin{table*}[t]
    \begin{minipage}[t]{0.7\linewidth}
        \centering
          \begin{adjustbox}{max width=\linewidth}
          \begin{tabular}{c|c|c|c|c|ccc}
            \toprule
            \multirow{2}{*}{Method} & Num & Trainable & GPU & \multirow{2}{*}{FLOPs} & \multicolumn{3}{c}{\textbf{100-10} (6 tasks)} \\
              & Prompts & Params & Memory &  & \textit{1-100} & \textit{101-150} & \textit{all} \\
            \midrule
            \multirow{3}{*}{\shortstack{CoM-\\Former~\cite{(comformer)cermelli2023comformer}}} & 100 & 44.38M & 3.28G & ~~97.46G & 36.0 & 17.1 & 29.7 \\
             & 150 & 44.40M & 3.66G & ~~99.35G & 36.5 & 15.9 & 29.6 \\
             & 200 & 44.43M & 4.03G & 101.27G & 36.7 & 12.1 & 28.5 \\
            \midrule
            \multirow{4}{*}{\textbf{ECLIPSE}} & 100 \footnotesize{~~(=50+10$\times$5)}  & ~~0.55M & 0.59G & ~~97.43G & 41.2 & 18.7 & 33.7 \\
                                  & 150 \footnotesize{(=100+10$\times$5)} & ~~0.55M & 0.59G & ~~99.27G & \textbf{41.4} & \textbf{18.8} & \textbf{33.9} \\
                                  & 200 \footnotesize{(=100+20$\times$5)} & ~~0.57M & 0.66G & 101.14G & 41.0 & 18.7 & 33.4 \\
                                  & 300 \footnotesize{(=100+40$\times$5)} & ~~0.62M & 0.79G & 104.83G & 39.6 & 18.4 & 32.5 \\
            \bottomrule
          \end{tabular}
          \end{adjustbox}
          \caption{
            \textbf{Effect of the number of prompts} and resulting \textbf{computational complexity}.
            We measure the GPU memory per a single training image.
            The number of prompts in ECLIPSE is denoted as \footnotesize{\texttt{(NUM{~}BASE{~}PROMPTS)}{+}\texttt{(NUM{~}NEW{~}PROMPTS)}{$\times$}\texttt{(NUM{~}STEPS)}}.
          }
          \label{tab:analysis_num_prompts}
          \vspace{-3mm}
    \end{minipage}
    \hspace{6mm}
    \begin{minipage}[t]{0.25\linewidth}
          \begin{adjustbox}{max width=\linewidth}
          \begin{tabular}{c|ccc}
            \toprule
            \multirow{2}{*}{$\delta$} & \multicolumn{3}{c}{\textbf{100-10} (6 tasks)} \\
               & \textit{1-100} & \textit{101-150} & \textit{all} \\
            \midrule
            0.3 & 39.5 & 14.2 & 31.0 \\
            0.4 & 40.9 & 17.0 & 32.9 \\
            \textbf{0.5} & \textbf{41.4} & \textbf{18.8} & \textbf{33.9} \\
            0.6 & 40.1 & 18.3 & 32.8 \\
            0.7 & 38.4 & 15.3 & 30.7 \\
            \bottomrule
          \end{tabular}
          \end{adjustbox}
          \caption{
            \textbf{Effect of $\delta$} that is the post-processing hyperparameter for modulation of logit manipulation.
          }
          \label{tab:analysis_delta}
          \vspace{-3mm}
    \end{minipage}
\end{table*}

\begin{table}[t]
  \centering
      \begin{adjustbox}{max width=0.9\linewidth}
      \begin{tabular}{c|c|c|ccc}
        \toprule
        Prompt & Deep & Logit & \multicolumn{3}{c}{\textbf{100-10} (6 tasks)} \\
        Tuning & Prompt & Mani  & \textit{1-100} & \textit{101-150} & \textit{all} \\
        \midrule
        &  &  & ~~0.2 & ~~3.9 & ~~1.3 \\
        \checkmark &  &  & 11.1 & ~~3.9 & ~~8.7 \\
        \checkmark & \checkmark &  & ~~8.2 & 15.0 & 12.8 \\
        \checkmark &  & \checkmark & 40.5 & 14.0 & 31.7 \\
        \checkmark & \checkmark & \checkmark & 41.4 & 18.8 & 33.9 \\
        \bottomrule
      \end{tabular}
      \end{adjustbox}
      \caption{
        \textbf{Effect of the proposed components}.
      }
      \label{tab:analysis_logit_mani}
\end{table}

\subsection{Experimental Results}
\begin{figure*}[t]
    \centering
    \includegraphics[width=1\linewidth]{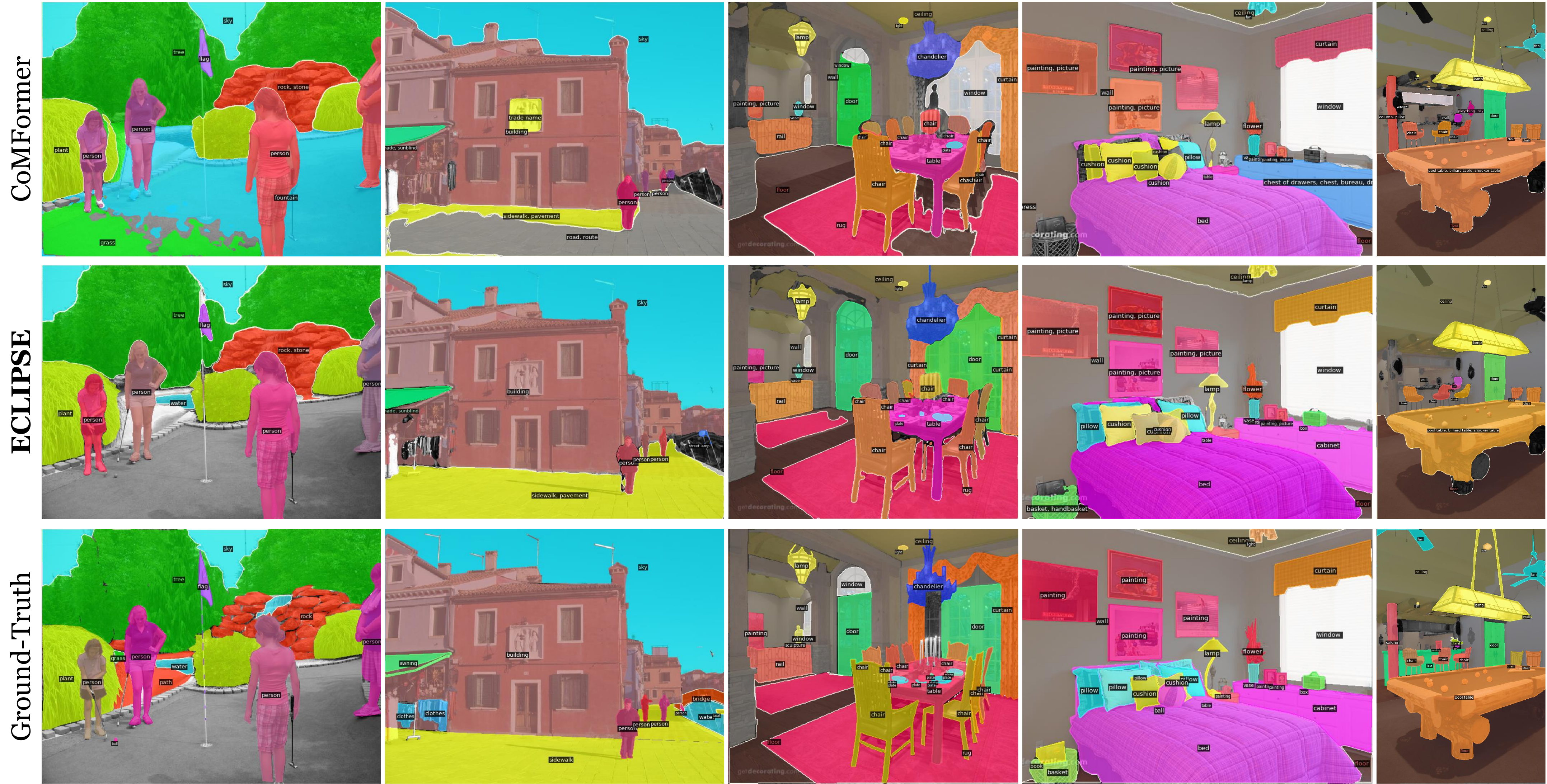}
    \caption{\textbf{Qualitative comparisons between ECLIPSE and CoMFormer~\cite{(comformer)cermelli2023comformer}} on the ADE20K \texttt{100-10} continual panoptic segmentation scenario.
    Our ECLIPSE shows more robust results against catastrophic forgetting without reliance on distillation strategies.
    }
    \label{fig:qualitative_result}
    \vspace{-2mm}
\end{figure*}

\paragraph{Continual Panoptic Segmentation.}
We evaluate our approach against three previous methods ($i.e.,$ MiB~\cite{(MiB)cermelli2020modeling}, PLOP~\cite{(PLOP)douillard2021plop}, and CoMFormer~\cite{(comformer)cermelli2023comformer}) and the basic fine-tuning (FT) approach, all using the Mask2Former~\cite{(mask2former)cheng2022masked} network with ResNet-50 backbone.
We conduct experiments on six incremental scenarios and report their reproduced performances using the official implementation of CoMFormer.
The three previous methods depend on knowledge distillation and pseudo-labeling that involve more trainable parameters and doubled network forwarding.
Unlike them, our method stands out as the first without distillation, streamlining the continual learning process significantly.
As shown in Table \ref{tab:comparision_adps}, our method achieved a new state-of-the-art performance across all tested scenarios, requiring only 1.3\% of total trainable parameters.
Notably, our method demonstrates superior retention of previously learned knowledge, even with a large number of continual steps, as evidenced in scenarios such as \texttt{100-5} and \texttt{50-10}.
In addition, our visual prompt tuning approach with the proposed logit manipulation strategy effectively develops the plasticity of the model for new classes, even when the base knowledge of the model is diminished from 100 to 50 classes, as shown in Table \ref{tab:comparision_adps} (b).
Our superiority can also be found in the qualitative results in Figure \ref{fig:qualitative_result}.

\paragraph{Continual Semantic Segmentation.}
We extend our evaluation to the semantic segmentation benchmark and compare our method against six previous methods~\cite{(SDR)michieli2021continual,(UCD)yang2022uncertainty,(SPPA)lin2022continual,(RC)zhang2022representation,(ssul)cha2021ssul,(REMINDER)phan2022class} using DeepLab-V3~\cite{(deeplabv3)chen2017rethinking} network and three methods (MiB, PLOP, and CoMFormer) using Mask2Former network, all using the same ResNet-101 backbone.
Here again, our method is the only one without distillation.
As shown in Table \ref{tab:comparision_adss_100}, our method achieves the best trade-off between catastrophic forgetting and plasticity, especially in difficult scenarios like \texttt{100-5}.
Although Mask2Former-based methods show slightly higher mIoU scores of new classes than ours in the \texttt{100-50} scenario, the reason is that our method focuses more on catastrophic forgetting than plasticity and their distillation strategy is more effective in easier scenarios.
Furthermore, compared to SSUL~\cite{(ssul)cha2021ssul} that employs model freezing and fine-tunes new classifier layers and enhances plasticity using off-the-shelf saliency maps, our approach outperforms without using saliency maps.

\section{Analysis}
In this section, we delve into the details of our method using the ADE20K panoptic segmentation \texttt{100-10} scenario.

\paragraph{Effect of the number of queries.}
In Table \ref{tab:analysis_num_prompts}, we conduct experiments to analyze the effect of the number of prompts in continual panoptic segmentation.
The total number of prompts $N$ is defined as $N{=}B{+}C{\times}T$, where $B$ is the number of base prompts (step $1$), $C$ is the number of incremented prompts (step $t{>}1$), and $T$ is the total continual steps.
By default, we set $B{=}|\mathcal{C}^{t}|$ and $C{=}|\mathcal{C}^{t}|$, that is, $N{=}150{=}100{+}10{\times}5$ for \texttt{100-10} scenario.
When we use the same number of prompts as our baseline work, CoMFormer~\cite{(comformer)cermelli2023comformer}, our ECLIPSE still significantly outperforms them.
We also found that increasing the number of prompts in ECLIPSE does not help in improving the performance because the more incremented prompts lead to more false-positive predictions due to over-weighting to new classes; this tendency also appears in CoMFormer.
Conversely, decreasing the number of prompts from 150 to 100 can save the FLOPs but marginally drop the performance by 0.2\% due to the reduced capacity of the model.

\begin{figure}[t]
    \centering
    \includegraphics[width=\linewidth]{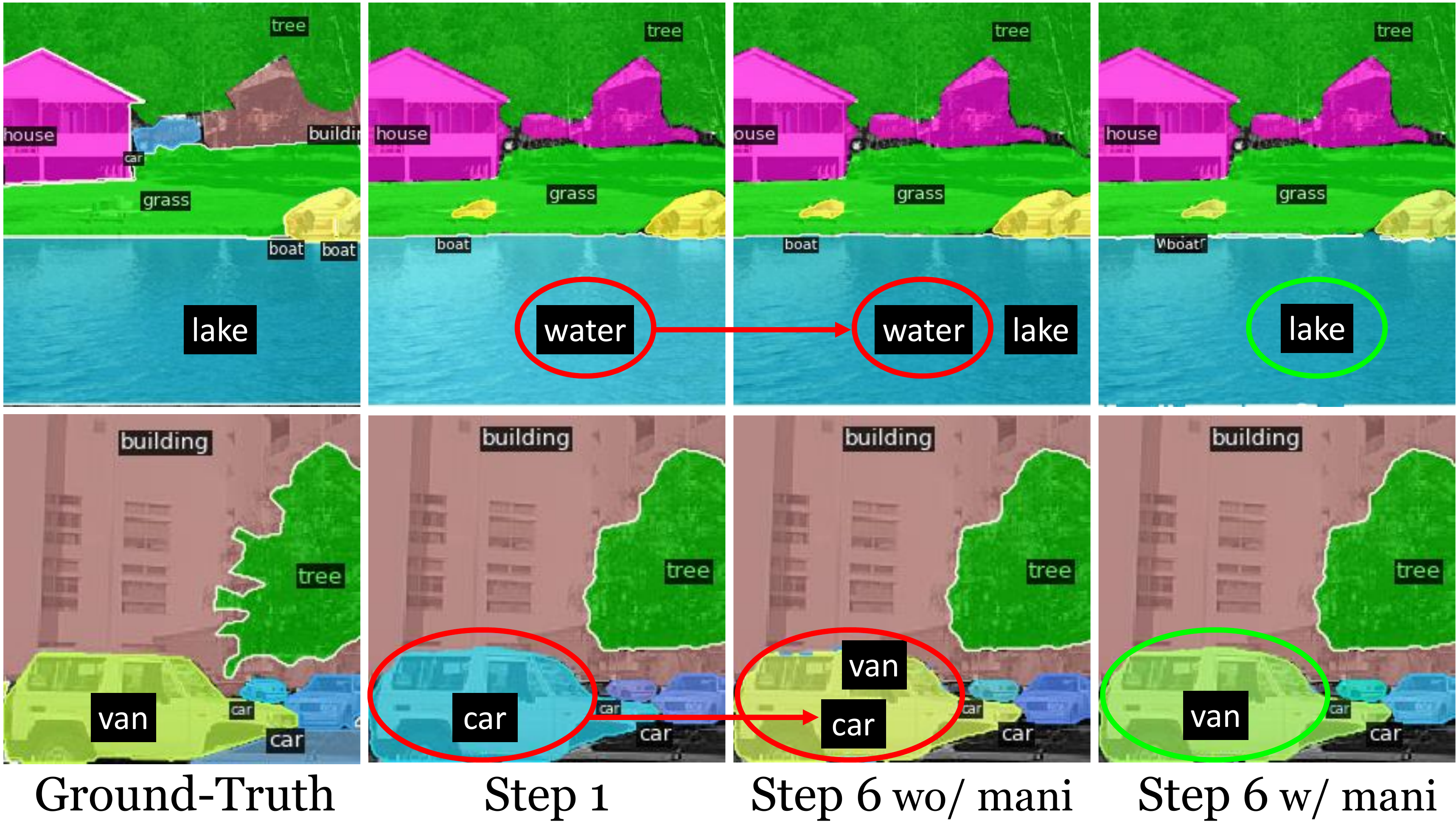}
    \caption{\textbf{Qualitative samples for logit manipulation.} At step 1, the model, which learned classes $\mathcal{C}^{1}$ containing \texttt{water} and \texttt{car}, can produce incorrect predictions due to semantic confusion with unexplored classes; these errors propagate forward continuously, resulting in overlapping predictions for one object (3rd column).
    After the model learns new classes containing \texttt{lake} and \texttt{van} at step 6, the logit manipulation can suppress the prior errors.
    }
    \label{fig:logit_manu_sample}
\end{figure}

\paragraph{Computational complexity.}
In Table \ref{tab:analysis_num_prompts}, we measure the computational complexity of ECLIPSE and CoMFormer according to the number of prompts.
Compared to CoMFormer, our ECLIPSE, which obviates the need for distillation strategies, shows 80 times less trainable parameters and 5.6 times less GPU memory usage for training.
In addition, CoMFormer processes all prompts together, leading to complexity that grows quadratically with the number of queries, $O(N^{2})$.
In contrast, our ECLIPSE processes each set of queries $\mathbf{Q}^{t}$ separately, multiplying the complexity by the number of steps, $O(B^{2}){+}O(C^{2}){\times}T$, where $N{=}B{+}C{\times}T$.
Even if the total number of queries is the same ($e.g., N{=}100)$, our method shows slightly advanced computational complexity (97.43G $v.s.$ 97.46G).
However, we argue that increasing the scalability of the model as the number of classes gets larger is the natural step even in fully supervised models and our increased computation due to new prompts is marginal compared to the total FLOPs of the model (3.8\% FLOPs increasing by 100 additional prompts).

\paragraph{Effect of visual prompt tuning.}
To validate the impact of the prompt tuning, we skip the model freezing and fine-tune all parameters of the model including new prompt sets without distillation strategies.
As shown in the last row of Table \ref{tab:analysis_logit_mani}, the performance is extremely dropped because the model substantially suffers from catastrophic forgetting due to the absence of model freezing or distillation strategies.

Moreover, as mentioned in Section \ref{method:prompt_tuning}, we have two prompt tuning strategies, termed \textit{shallow} and \textit{deep}.
We adopt the \textit{deep} strategy by default and the \textit{deep} requires 100K more trainable parameters than the \textit{shallow}.
The result in the third row of Table \ref{tab:analysis_logit_mani} shows that adopting the \textit{shallow} strategy noticeably drops the PQ for new classes (18.8\%{$\rightarrow$}14.0\%) due to the reduced plasticity of the model.
Considering the total number of model parameters is 63.4M, the 100K additional parameters in the \textit{deep} strategy are efficient in developing the plasticity of the model.

\vspace{-2mm}
\paragraph{Effect of the logit manipulation.}
To analyze the effect of the logit manipulation, we conduct an ablation study, as shown in the second row of Table \ref{tab:analysis_logit_mani}.
Without the logit manipulation, the prior errors caused by semantic confusion propagate forward (Figure \ref{fig:logit_manu_sample}), and the definition of \texttt{no-obj} continuously shifts as continual learning progresses, extremely diminishing the performance of old classes.

In addition, we have a post-processing hyperparameter $\delta$ for the logit manipulation.
Table \ref{tab:analysis_delta} shows that $\delta$ of 0.5 is a suitable value.
We note that since the $\delta$ is a post-processing hyperparameter, it requires much less endeavor for tuning. 

\paragraph{Exploring advanced frozen parameters.} 
To demonstrate the potential for further improving ECLIPSE, we explore the impact of using more advanced frozen parameters of the base model.
When employing the more powerful backbone network, Swin-L~\cite{(swin)liu2021swin}, we observe significant improvements in all tested scenarios, as shown in Table \ref{tab:comparision_adps}.
Moreover, leveraging pre-trained weights from other datasets ($e.g.,$ COCO~\cite{(coco)lin2014microsoft}) substantially boosts the performance of ours, as shown in our supplementary material.

\section{Conclusion and Future Direction}

We presented a groundbreaking method in the field of continual panoptic segmentation.
By integrating VPT with our innovative logit manipulation technique, we effectively addressed key challenges in the field. 
Our method not only ensures the preservation of previously learned information but also adapts to new class information.
The experimental results on ADE20K dataset highlight the superiority of our approach, achieving state-of-the-art performance with a notable reduction in the number of trainable parameters. 
Our future direction may involve optimizing increased computational complexity resulting from expanding prompt sets, particularly when dealing with a massive number of classes.

\vspace{1mm}
\noindent{\textbf{Acknowledgment.}} This work was supported by Institute for Information {\&} communications Technology Promotion(IITP) grant funded by the Korea government(MSIT) (No.2019-0-00075 Artificial Intelligence Graduate School Program(KAIST).

\nocite{(max-deeplab)wang2021max}
\nocite{(panoptic-segformer)li2022panoptic}
\nocite{(ILT)michieli2019incremental}
\nocite{(MTN)gu2021class}
\nocite{(MMA)cermelli2022modeling}
\nocite{(CL4WSIS)hsieh2023class}
\nocite{kirkpatrick2017overcoming}
\nocite{douillard2022dytox}
\nocite{li2017learning}
\nocite{maracani2021recall}
\nocite{vaswani2017attention}
\nocite{kim2023devil}
\nocite{kim2022beyond}
\nocite{kim2021discriminative}

\newpage
{
    \small
    \bibliographystyle{ieeenat_fullname}
    \bibliography{ms}
}

\clearpage

\onecolumn
\appendix
\renewcommand{\thesection}{\Alph{section}}
\section*{Appendix}

\section{Additional Analysis}

\subsection{Impact of Class Ordering on Performance}
\begin{figure*}[h]
    \centering
    \includegraphics[width=0.8\linewidth]{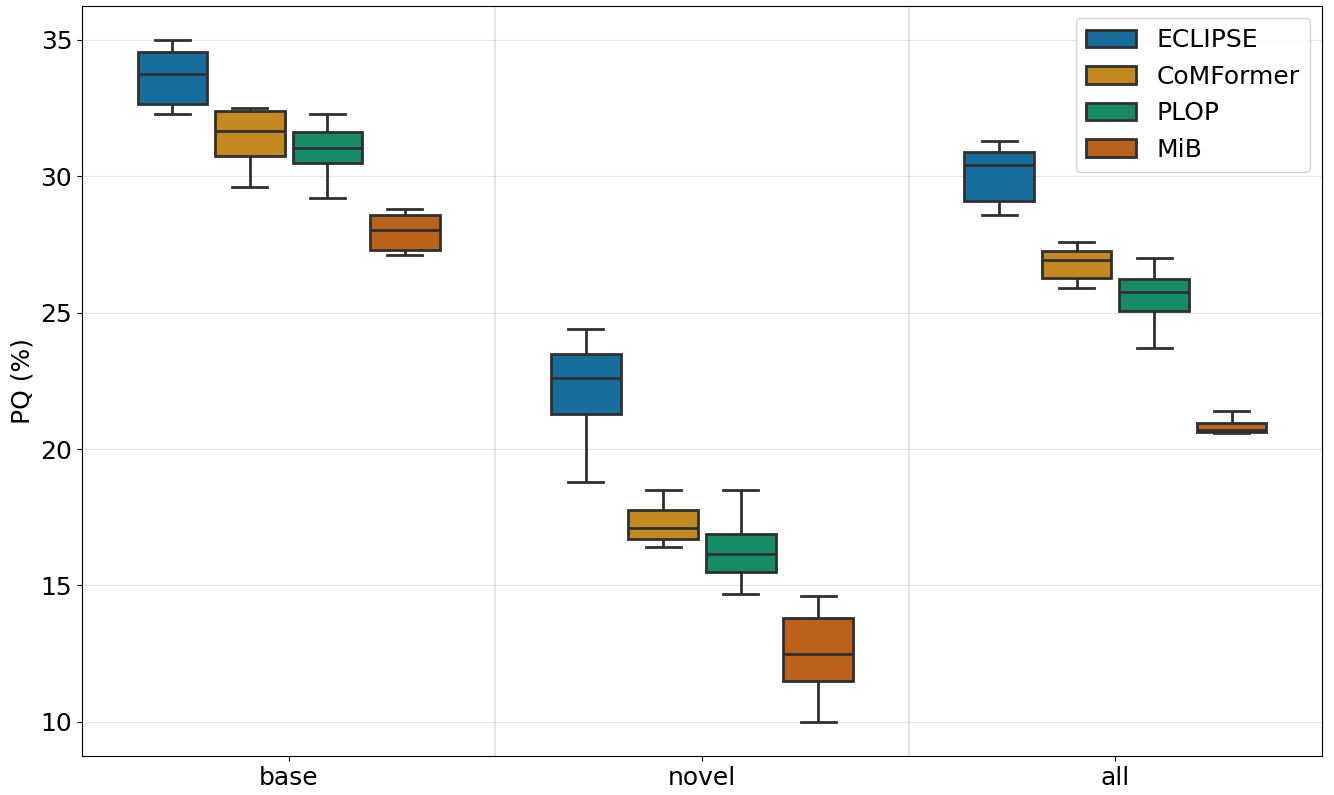}
    \caption{PQ distributions for 10 different class-orderings in the ADE20K panoptic segmentation \texttt{100-10} scenario.}
    \label{fig:class_ordering}
\end{figure*}

We delve deeper into the robustness of our method concerning class ordering. 
We carry out experiments on the ADE20K panoptic segmentation \texttt{100-10} scenario, employing 10 different class orderings. 
Note that we randomly shuffle the base class order 10 times to generate these varied orderings. 
In Figure~\ref{fig:class_ordering}, PQ distributions are illustrated through boxplots. 
Remarkably, our ECLIPSE demonstrates resilience to changes in class ordering, consistently outperforming other methods.

\subsection{Continual Panoptic Segmentation under the Disjoint Setting}
The seminal work~\cite{(MiB)cermelli2020modeling} introduced two different settings, \textit{disjoint} and \textit{overlap}. 
Since the overlap setting is more challenging and realistic, we mainly followed it in our main paper.
Here, we provide the experimental results on ADE20K~\cite{(ade20k)zhou2017scene} continual panoptic segmentation under the \textit{disjoint} setting.
Table \ref{tab:comparision_adps_disjoint} shows the superiority of ECLIPSE compared to existing continual panoptic segmentation methods.

\begin{table*}[h]
  \centering
  \begin{adjustbox}{max width=\linewidth}
  \begin{tabular}{c|c|c|c|ccc|ccc|ccc}
    \toprule
    \multirow{2}{*}{Method} & \multirow{2}{*}{Backbone} & Trainable & \multirow{2}{*}{KD} & \multicolumn{3}{c|}{\textbf{100-5} (11 tasks)} & \multicolumn{3}{c|}{\textbf{100-10} (6 tasks)} & \multicolumn{3}{c}{\textbf{100-50} (2 tasks)} \\
     & & Params & & \textit{1-100} & \textit{101-150} & \textit{all} & \textit{1-100} & \textit{101-150} & \textit{all} & \textit{1-100} & \textit{101-150} & \textit{all} \\
    \midrule
    MiB~\cite{(MiB)cermelli2020modeling} & R50 & 44.9M & \checkmark & 20.5 & 4.3 & 15.1 & 27.7 & 7.1 & 20.8 & 33.7 & 10.5 & 26.0 \\
    PLOP~\cite{(PLOP)douillard2021plop} & R50 & 44.9M & \checkmark & 19.2 & 8.8 & 15.8 & 28.9 & 10.6 & 22.8 & 34.8 & 12.4 & 27.4 \\
    CoMFormer~\cite{(comformer)cermelli2023comformer} & R50 & 44.9M & \checkmark & 20.1 & 8.2 & 16.1 & 29.7 & 10.3 & 23.3 & 34.7 & 13.2 & 27.6 \\
    \midrule
    \textbf{ECLIPSE} & R50 & \textbf{0.60M} & & 34.4 & 8.9 & 25.9 & 34.4 & 10.2 & 26.4 & 35.2 & 13.3 & 27.9 \\
    \bottomrule
  \end{tabular}
  \end{adjustbox}
  \caption{
    \textbf{Continual Panoptic Segmentation} results on ADE20K dataset in PQ under the \textit{disjoint} setting. \textit{KD} denotes using distillation strategies, which demands more trainable parameters and computational overhead. All methods use the same network of Mask2Former~\cite{(mask2former)cheng2022masked} with ResNet-50~\cite{(resnet)he2016deep} backbone. 
  }
  \label{tab:comparision_adps_disjoint}
\end{table*}

\subsection{Continual Panoptic Segmentation on COCO Dataset}
We validate our approach on the COCO panoptic segmentation benchmark~\cite{(coco)lin2014microsoft}, comprising 100K training and 5K validation images spread across 133 classes. 
For the incremental protocol, we designate 83 base classes and increment by an additional 50 classes. 
We note that the class ordering of COCO panoptic segmentation consists of things and stuff in sequence. 
To conduct a more meaningful validation, we randomly shuffled this order: 

\noindent 
\texttt{\scriptsize{[1, 3, 10, 47, 58, 9, 88, 16, 126, 120, 17, 129, 35, 119, 59, 57, 54, 90, 75, 38, 80, 48, 131, 56, 95, 25, 43, 2, 68, 110, 32, 14, 29, 11, 7, 52, 83, 102, 84, 73, 5, 45, 117, 93, 87, 46, 118, 34, 61, 19, 77, 111, 63, 98, 130, 66, 79, 97, 33, 86, 127, 104, 64, 49, 36, 6, 91, 50, 112, 8, 65, 132, 92, 27, 122, 22, 51, 85, 115, 28, 89, 70, 62, 12, 101, 108, 125, 123, 39, 81, 20, 40, 41, 114, 128, 74, 18, 99, 100, 60, 30, 124, 69, 37, 13, 23, 116, 55, 26, 121, 71, 67, 106, 133, 42, 107, 105, 109, 82, 103, 76, 94, 24, 15, 78, 53, 21, 96, 72, 113, 44, 31, 4]}}.
\vspace{1mm}

\noindent Our method is compared against three baseline methods (MiB~\cite{(MiB)cermelli2020modeling}, PLOP~\cite{(PLOP)douillard2021plop}, and CoMFormer~\cite{(comformer)cermelli2023comformer}), all utilizing the ResNet-50 backbone network under the \textit{overlap} setting. 
As demonstrated in Table~\ref{tab:comparision_cocops}, our approach exhibits superior performance with considerably fewer trainable parameters.

\begin{table*}[h]
  \centering
  \begin{adjustbox}{max width=\linewidth}
  \begin{tabular}{c|c|c|c|ccc|ccc}
    \toprule
    \multirow{2}{*}{Method} & \multirow{2}{*}{Backbone} & Trainable & \multirow{2}{*}{KD} & \multicolumn{3}{c|}{\textbf{83-5} (11 tasks)} & \multicolumn{3}{c}{\textbf{83-10} (6 tasks)}  \\
     & & Params & & \textit{1-83} & \textit{84-133} & \textit{all} & \textit{1-83} & \textit{84-133} & \textit{all} \\
    \midrule
    MiB~\cite{(MiB)cermelli2020modeling} & R50 & 43.9M & \checkmark & 29.3 & 25.6 & 27.9 & 34.8 & 28.0 & 30.3 \\
    PLOP~\cite{(PLOP)douillard2021plop} & R50 & 43.9M & \checkmark & 34.0 & 27.1 & 31.4 & 37.7 & 31.1 & 35.2 \\
    CoMFormer~\cite{(comformer)cermelli2023comformer} & R50 & 43.9M & \checkmark & 34.2 & 27.3 & 31.6 & 37.7 & 31.5 & 35.4 \\
    \midrule
    \textbf{ECLIPSE} & R50 & \textbf{0.60M} &  & \textbf{36.9} & \textbf{31.7} & \textbf{34.9} & \textbf{38.1} & \textbf{34.5} & \textbf{36.7} \\
    \bottomrule
  \end{tabular}
  \end{adjustbox}
  \caption{
    \textbf{Continual Panoptic Segmentation} results on COCO~\cite{(coco)lin2014microsoft} panoptic segmentation dataset where the total number of classes is 133 in PQ under the \textit{overlap} setting. \textit{KD} denotes using distillation strategies, which demands more trainable parameters and computational overhead.
  }
  \label{tab:comparision_cocops}
\end{table*}

\subsection{Exploring Pre-trained Knowledge}
To demonstrate the potential for further improving ECLIPSE, we explore the impact of using more advanced frozen parameters of the base model.
We study the effect of the frozen parameters in continual segmentation using various pre-trained weights from Cityscape~\cite{(cityscape)cordts2016cityscapes}, Mapillary Vistas~\cite{(mapillary)neuhold2017mapillary}, and COCO~\cite{(coco)lin2014microsoft} panoptic segmentation datasets.
By default, we used the ImageNet pre-trained weights only for the backbone network (ResNet-50~\cite{(resnet)he2016deep}).
As shown in Table \ref{tab:analysis_pretraining}, using pre-trained weights on larger datasets ($e.g.,$ Cityscape{$\rightarrow$}Mapillary{$\rightarrow$}COCO) results in more noticeable performance improvements.
This result demonstrates the potential of our approach to further enjoy the more powerful expandability of the model.

\begin{table*}[h]
  \centering
  \begin{adjustbox}{max width=\linewidth}
  \begin{tabular}{c|ccc|ccc|ccc}
    \toprule
     \multirow{2}{*}{Pretrained} & \multicolumn{3}{c|}{\textbf{100-5} (11 tasks)} & \multicolumn{3}{c|}{\textbf{100-10} (6 tasks)} & \multicolumn{3}{c}{\textbf{100-50} (2 tasks)} \\
     & 1-100 & 101-150 & all & 1-100 & 101-150 & all & 1-100 & 101-150 & all  \\
    \midrule
    - & 41.1 & 16.6 & 32.9 & 41.4 & 18.8 & 33.9 & 41.7 & 23.5 & 35.6  \\
    Cityscape~\cite{(cityscape)cordts2016cityscapes} & 41.7 & 16.9 & 33.2 & 42.2 & 18.9 & 34.5 &  42.2 & 23.8 & 35.9 \\
    Mapillary~\cite{(mapillary)neuhold2017mapillary} & 42.5 & 17.2 & 34.0 & 42.9 & 19.8 & 35.2 &  43.0 & 24.1 & 36.3 \\
    COCO~\cite{(coco)lin2014microsoft} & 46.1 & 18.9 & 37.0 & 46.4 & 22.3 & 38.4 & 44.2 & 29.0 & 39.1 \\
    \bottomrule
  \end{tabular}
  \end{adjustbox}
  \caption{
    \textbf{Exploring pre-trained knowledge.}
    At the beginning of the continual learning process, we employ the pre-trained parameters to explore stronger frozen parameters of the base model.
    }
  \label{tab:analysis_pretraining}
\end{table*}

\end{document}